\newif\ifarxiv\arxivtrue

\ifarxiv
  \documentclass[doc,floatsintext]{apa7}
\else
  \documentclass[man,floatsintext,onecolumn]{apa7}
  \usepackage{fullpage}
\fi
\usepackage{booktabs}

\usepackage{natbib}
\usepackage{microtype}
\usepackage{graphicx}
\usepackage{subfigure}
\usepackage{setspace}
\usepackage{centernot}
\usepackage{cancel}
\usepackage{booktabs}
\usepackage{hyperref}
\usepackage{dashrule}

\usepackage{algorithm}
\usepackage{algpseudocode}
\usepackage{cancel}
\usepackage{amsmath}
\usepackage{amssymb}
\usepackage{mathtools}
\usepackage{amsthm}
\usepackage{scalerel}
\usepackage{enumitem}
\usepackage{newpxtext,newpxmath} %

\usepackage[capitalize,noabbrev,nameinlink]{cleveref}
\creflabelformat{equation}{#2#1#3}
\usepackage{csquotes}
\usepackage{multirow}
\usepackage{mathrsfs}
\usepackage{mdframed}
\usepackage{tcolorbox}
\tcbuselibrary{breakable} %
\tcbuselibrary{skins}  %
\usepackage{bm}
\usepackage{bbm,tipa}
\usepackage{mathbbol}

\DeclareSymbolFontAlphabet{\mathbb}{AMSb}
\DeclareSymbolFontAlphabet{\mathbbl}{bbold}

\usepackage{xcolor}
\definecolor{dred}{RGB}{153,80,43}
\definecolor{dblue}{RGB}{0,114,178}
\hypersetup{
  colorlinks=true,
  breaklinks=true,
  urlcolor=dblue, %
  linkcolor=dred, %
  citecolor=dblue
}

\usepackage{makecell} %

\newtheorem{theorem}{Theorem}
\newtheorem*{theorem*}{Theorem}

\newtcolorbox{innerproofbox}{
  breakable,    %
  enhanced,     %
  sharp corners,  %
  frame hidden,   %
  borderline west={0.5pt}{0pt}{black, line cap=butt}, %
  colback=white,  %
  before skip=0pt, %
  boxsep=0pt,    %
  top=2mm,     %
  bottom=1mm,    %
  left=8pt,     %
  right=0pt,    %
}

\usepackage{etoolbox}
\makeatletter
\patchcmd{\NAT@test}{\else \NAT@nm}{\else \NAT@hyper@{\NAT@nm}}{}{}
\makeatother

\newcommand{\creflink}[1]{\hyperref[#1]{\textcolor{blue}{\cref{#1}}}}

\ifarxiv
\fi

\title{The syntax and semantics of goals}
\shorttitle{The syntax and semantics of goals}

\authorsnames{David M. Abel$^1$, Mark K. Ho$^2$}
\authorsaffiliations{\vspace{2ex}
$^1$Department of Informatics\\University of Edinburgh\\\texttt{david.abel@ed.ac.uk}\\[2ex]
$^2$Department of Psychology\\New York University\\\texttt{mark.ho@nyu.edu}}


\ifarxiv
\authornote{To appear in a special issue of \emph{Topics in Cognitive Science} on Goal-Centric Perspectives in Cognitive Science.}

\else
\note{\vspace{14mm}
\textbf{Word Count:} 1. Text : 5,249 | 2. Abstract: 194 | 3. Table/Figure Captions: 161 | 4. Appendix: N/A | 5. Notes and Acknowledgments: 19}
\fi

\abstract{In both cognitive science and computer science, goals are conceptualized as cognitive states that flexibly combine with world knowledge to organize and specify purposeful behavior. In this way, \emph{goals are compositional representations whose content relates to rational behavior}. We here draw attention to goals as representations and their content because it highlights a parallel with other areas in cognitive science---in particular, the syntax-semantics interface in linguistics and logic---while also foregrounding foundational questions about the \emph{expressivity}, \emph{design}, and \emph{efficiency} of different goal representations. For example, goals are typically taken as fixed and imposing constraints on desirable behaviors, but we can also identify constraints on goal representations themselves, such as whether a particular goal language is sufficiently expressive to capture behaviors of interest, or whether different goal representations capture the same behavior. Here, we synthesize work that aims to characterize the properties of different goal representations and suggest these are points of a broader design space. We close by discussing how distinguishing the form and meaning of goals can elucidate the implicit assumptions we make about goals, inform the study of interactions between higher-level cognition and motivation, and isolate axes of variation for different conceptions of goals.}

\begin{document}
\maketitle
\ifarxiv\setlength{\parindent}{1.5em}\fi %

\pagebreak

\section{Introduction}

Goal representations are central to the study of biological intelligences~\citep{ball2023organisms} and the design of artificial intelligences~\citep{russell2010artificial}. Indeed, our intuitions about goals underpin much of our everyday theory of mind~\citep{premack1978does}. But, what are goals? What distinguishes goals from other psychological states, such as reflexes, habits, or knowledge about the world? How do goals interact with these other representations? And if goals are a kind of representation, what exactly do they represent?

To address these questions, we begin by highlighting two key properties of goal representations: first, \emph{goals control rational behavior}; second, \emph{goals are compositional}. 
The first property, that goals control rational behavior, simply means that goals are control representations that relate to behavior via a principle of rationality~\citep{miller2001integrative,cohen2017cognitive,dennett1989intentional}. As an example, consider the goal to \emph{run the Boston marathon}. To successfully run the Boston marathon, one can do several things to make that more likely, including: devise a training regimen, eat healthy, not skip practice, and show up at the starting line on the day of the marathon. Having a goal like running the Boston marathon is to have a mental state that orients behaviors towards those likely to achieve the goal and away from those unlikely to achieve the goal; in other words, goals orient an intelligent agent's behavior towards instrumentally rational, teleological action~\citep{dennett1989intentional,bratman1987intention,dickinson2000causal}. This capacity to rationally control or override otherwise non-rational behavior distinguishes goals from other representations, such as desires, which may not affect behavior, or habits, which may not be rational~\citep{berridge2003parsing,wood2016psychology}. For this reason, goals are viewed as central to processes that override or flexibly configure behavior (or even other cognitive processes) in a rational, task-dependent manner. This includes processes such as planning~\citep{mattar2022planning}, meta-cognition~\citep{griffiths2019doing}, social cognition~\citep{jara2016naive}, self-regulation~\citep{latham1991self}, cognitive control~\citep{cohen2017cognitive}, and mental effort~\citep{shenhav2017toward}.

The second property, that \emph{goals are compositional}, means that goal representations combine with other representations in a manner that systematically relates to behavior~\citep{velez2017interpreting,davidson2024toward,bakermans2025constructing}. For example, given two simple goals, \emph{run the Boston marathon} and \emph{avoid injury}, one can combine them into a composite goal, \emph{run the Boston marathon without getting injured}. This new goal specifies behavior in ways that are systematically similar to and different from those specified by the original goal---e.g., one will still go for practice runs but only gradually increase the length of practice runs to avoid injury. Moreover, not only do goals compose with other goals, they flexibly compose with new knowledge about the world. For instance, if one wanted to avoid injury and then learned about the importance of stretching consistently for that goal, one would change their behavior to incorporate stretching into their practice routine. Indeed, the property that goals can compose with world knowledge to flexibly guide behavior is perhaps \emph{the} hallmark of goal-directed behavior. Experimental manipulations such as \emph{reward devaluation}, in which a previously rewarding outcome stops being rewarding (e.g., moving water in a maze to a new location), explicitly leverage this property to assess whether a behavior is goal-directed or simply habitual (e.g., if a rat stops navigating to the former location of the water given minimal training, this is taken as evidence for a goal representation)~\citep{dickinson1985actions}.

\begin{figure}[]
\centering
\includegraphics[width=\linewidth]{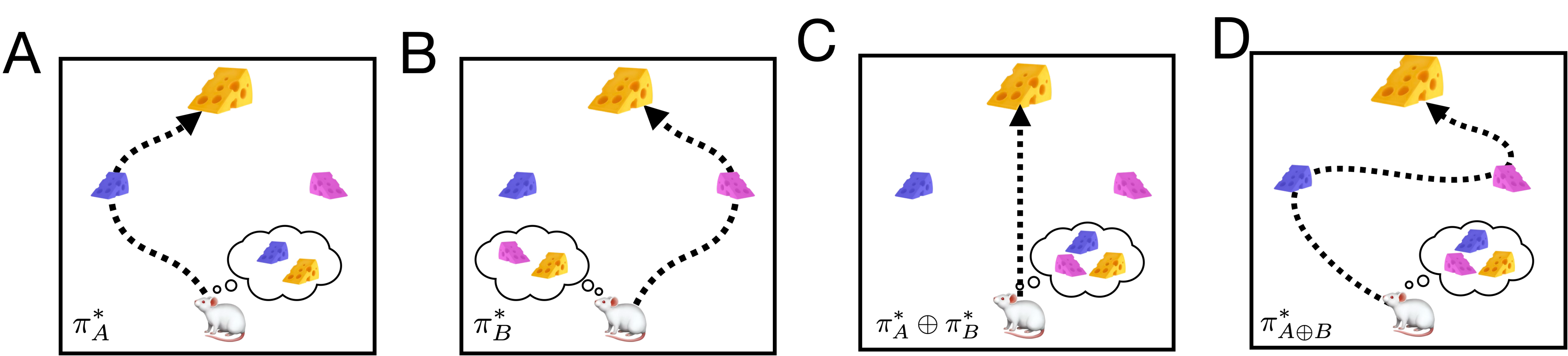}
    \caption{Behavioral composition is not rationality-preserving. A rat could have the goal of (A) reaching the big cheese and the small blue cheese or (B) the big cheese and the small pink cheese. The optimal behaviors are represented as dotted lines and denoted $\pi^*_A$ and $\pi^*_B$, respectively. (C) Directly composing the \emph{behaviors} from A and B (e.g., $\pi^*_A \oplus \pi^*_B$) might have the rat attempt to walk straight to the cheese. (D) This is different from first composing the \emph{goals} of A and B to derive a compound goal---reach the big cheese and both small cheeses---and then acting rationally with respect to the compound goal (i.e., $\pi^*_A \oplus \pi^*_B \neq \pi^*_{A \oplus B}$).}
    \label{fig:compositionality}
\end{figure}

In short, goals control rational behavior and goals are compositional.\footnote{To be clear, these two properties are not intended to be taken as necessary and sufficient conditions of goals, but rather, in our view, motivate the design and analysis of most formal accounts of goals.} However, we argue that the precise way in which goals and behaviors compose and relate to one another is far from obvious.
Consider the example shown in figure~\ref{fig:compositionality}, in which a rat could be in two situations: one in which they have the goal of reaching the big yellow cheese and the small blue cheese (goal A), and, separately, in which they have the goal of reaching the big yellow cheese and the small pink cheese (goal B). A and B each produce different goal-directed behaviors, which we can call behaviors A and B. Notice, however, that while we could directly compose behaviors A and B by, say, averaging (panel C), this is not the same as the rational behavior for the composed goal $A \oplus B$---i.e., reach the large cheese while collecting \emph{both} small cheeses (panel D). This example illustrates a general problem: Even though goals control rational behavior and are compositional, \emph{direct composition of rational behavior does not preserve rationality}.

\subsection{The syntax-semantics interface}

The preceding discussion highlights a tension between goals as specifications of rational behavior and goals as representations that systematically compose. We will not fully resolve this tension in this paper. Rather, our aim is to provide a lens for clarifying the core issues and frame recent research from this perspective. To this end, our primary claim is that the study of goals should borrow an analytical distinction developed in another area of cognitive science: the syntax-semantics interface from linguistics and logic~\citep{montague1970english,jacobson2014compositional}.

In the study of natural and artificial languages, semantics focuses on the \textit{meaning} of expressions. For example, the word ``Rex'' can refer to a dog while the phrase ``the neighbor's cat'' can refer to a particular cat. In contrast, syntax focuses on how simpler linguistic elements combine to form complex expressions and the rules for combining them, also called a \emph{grammar}. For example, in English, ``Rex chased the neighbor's cat'' is a well-formed expression consisting of the words ``Rex'', ``chased'', ``the'', ``neighbor's'', and ``cat'', whereas ``chased Rex the neighbor's cat'' is not. The syntax-semantics interface concerns how the form of expressions interacts with the meaning of expressions. In particular, given a language, one can ask about its \emph{compositional semantics}, that is, how the semantic meaning of complex expressions derives from the meaning of its simpler constituents and the syntactic rules used to combine them~\citep{jacobson2014compositional}. The utterance ``Rex chased the neighbor's cat'', for instance, denotes a state of affairs in which a dog named Rex chased a cat, and its meaning is derived from the meanings of its constituents (i.e., ``Rex'', ``chased'', and ``the neighbor's cat'') and their composition (i.e., a subject-verb-object construction).

A complete review of the syntax-semantics distinction and its history is beyond our scope, but it is worth outlining some ways in which it serves as a valuable analytic tool in the study of language. First, the distinction between the syntax and semantics of a language allows us to talk precisely about \emph{semantic equivalence}. For example, in English, ``Rex chased the neighbor's cat'' means effectively the same thing as ``The neighbor's cat was chased by Rex'' because they denote the same state of affairs. The difference between these two statements is their syntax: the first statement uses an active construction, while the second statement uses a passive construction~\citep{siewierska1984passive}. Second, we can ask about the \emph{expressivity} of the lexicon and syntax of a language with respect to a semantics. Different languages provide different grammatical forms that can make it easier or harder to express certain meanings. For example, even though all languages can express the source of information, some language families (e.g., Turkic languages) have a distinct grammatical form for denoting how some fact was acquired (e.g., by direct experience or from another's testimony) that must always be specified~\citep{johanson2003evidentiality}. Finally, a compositional semantics provides a framework for characterizing systematic patterns of \emph{learning}. As an example, research on early word learning suggests that infants acquire nouns before verbs because nouns tend to denote objects whereas verbs tend to denote actions and relations among objects~\citep{golinkoff2008toddlers}. Nouns and verbs are syntactic notions while objects and relations between them are semantic ones.

Beyond linguistics, the relationship between the syntactic and semantic features of a representational system is a recurring theme throughout cognitive science. For instance, compositionality has been argued to be an important general principle not only of language, but cognition and learning more broadly~\citep{fodor1988connectionism,lake2017building,kirby2015compression}. Similarly, within computer science, delineating interactions between syntax and semantics is central to fields such as logic, programming languages, compilers, and formal verification~\citep{pierce2002types}. Here, we propose that the syntax-semantic distinction can serve a similar role in organizing how we think about goals and agency.

\section{The syntax and semantics of goal representations}

How might the syntax-semantics distinction be applied to goals? Drawing on the analogy with language, any compositional semantics for goals needs to answer two questions: What is the semantic space being represented by goals? And what are the syntactic rules for composing goal representations?

We expect there to be many reasonable answers to these two questions that yield different implications. To ground our discussion, we start by focusing on how the syntax-semantics distinction might clarify issues in reinforcement learning, a formal framework for studying goal-directed behavior that is used in both cognitive science and artificial intelligence~\citep{sutton2018reinforcement,dayan2008reinforcement}. After taking a deep dive into this one ``point'' in the ``space'' of possible compositional semantics, we will discuss how this perspective can shed light on possible goal languages and other representations that interface with goals.

\subsection{The Expressivity of Goals in Reinforcement Learning}

In standard formulations of reinforcement learning, tasks are most often characterized as Markov Decision Processes (MDPs: \citeauthor{puterman2014markov}, \citeyear{puterman2014markov}). The commitment to the MDP as a mechanism for modeling tasks forefronts the two syntax-semantics questions directly: what are the semantics of goals in an MDP, and what are the syntactic rules for composing goals? Within an MDP, an agent's goal is represented by a reward function that assigns a number to each possible experience. As articulated by the reward hypothesis \citep{suttonwebRLhypothesis}, the agent's goal is then understood in terms of maximizing some accumulation of reward (typically, an expected discounted sum).

In this way, we have a clear syntactic picture: goals are \textit{represented} as rewards---numbers assigned to each experience---which can combine across time and possible consequences to form a composite value. However, this still does not provide a \textit{semantics} to these goals: what does a reward of "7" mean? What does the sequence of rewards "2, -1, 0, 4" mean? In other words, if we have two agents with two separate goals, how can we answer questions about the meaning of their respective reward sequences, or their reward functions more generally? To do so, we suggest it is necessary to interpret reward as a syntactic structure whose meaning is grounded in some other content---in this case, preferences over behavior.

In recent work, \cite{abel2021reward} make this study explicit by asking: what can reward functions of different kinds \textit{express}? In this way, they draw a relationship between the syntactic form of reward (its numerical value) and the meaning referred to by these values. Specifically, the arguments presented by \cite{abel2021reward} suggest that \textit{reward} is a communicative mechanism for the representational content the reward refers to: a preference relation on policies.
That is, Abel et al. consider (among several choices) a simple class of preference relations over policies they call a Set Of Acceptable Policies (SOAPs). A SOAP is a preference relation on the class of deterministic functions that assign action to state. By definition a SOAP only induces two equivalence classes: the good strategies for acting, and the bad. Abel et al. ask: Can Markov reward functions \textit{express} each possible SOAP? A Markov reward function is so called as it only depends on the most recent state experienced, as opposed to the full history of interaction. 
Implicitly, this question is one that exploits the syntax and semantics of goals---can reward, a syntactic structure, contain the necessary representational content of a particular kind of preference ordering? Thus, it is a question about the expressive power of the \textit{syntactic language of reward} for referring to the \textit{semantic language of preferences}.

Crucially, without further commitments, a preference relation under-determines a reward function. It is similar in spirit to a set of logical predicates without their connectives, or a choice of which logical rules to embrace: we have not yet made a commitment for \textit{how} the reward needs to \textit{compose} in order to form a value function that actually does the ordering.
In logic, we might reject the law of excluded middle as famously argued by \cite{priest1983logical}, or \textit{Modus Ponens} as explored by \cite{lewis1895tortoise} and \cite{decker2006modality}, or choose to embrace real-numbered truth values as in fuzzy logic \citep{zadeh1988fuzzy}. Such choices can impact the way we can compose our logical primitives into sentences, or the truth value of those corresponding sentences.
In an MDP, even value is not semantic on its own; assigning a specific number to some state of affairs, such as ``4.3", carries no extrinsic meaning. This is in part due to a key aspect of the decision theoretic foundations that underlie much of consequentialist views of decision making. Specifically, the classical work by von Neumann and Morgenstern (vNM: \citeauthor{vonneumann1953theory}, \citeyear{vonneumann1953theory}) establishes that any utility function $u$ that captures a vNM-satisfying preference is \textit{unique up to affine transformations}. In this way, we could select two distinct utility functions $u$ and $u'$ that realize the exact same preference, but assign arbitrarily different utilities to an individual state of affairs. As a result, the precise number associated with value only takes on meaning in so far as it refers to a preference relation. Two utility functions can thus \textit{mean} the same thing when they refer to the same preference ordering. The same argument holds of reward, thus solidifying their status as syntactic aspects of goals.

In goal space, \cite{abel2021reward} explore this perspective in two steps. First, they turn reward into the standard compositional view of \textit{expected cumulative discounted reward}, for a choice of geometric discount factor. 
A value function orders the states of the Markov process underlying the MDP. To connect the preferences implicit in SOAPs with these value functions, Abel et al. again take a standard view to treat the \textit{expected cumulative discounted reward} of each specific policy in the SOAP, evaluated at the start-state of the MDP. Then, a reward function can \textit{express} a SOAP just when the start-state value of every good policy is strictly higher than the start-state value of every bad policy. 
In other words: if a SOAP is a splitting of the decision-making strategies into "good" and "bad", can the start-state value function induced by a reward function define the boundary separating good from bad? Together, these commitments yield the following result about the expressivity of Markov reward.

\begin{mdframed}
\begin{theorem}[Informal, Theorem 4.1 of \cite{abel2021reward}]
    For any finite Markov controlled process, there exists a SOAP such that there is no Markov reward function that expresses the SOAP.
\end{theorem}
\vspace{2mm}
\end{mdframed}

\noindent This result reveals a limitation to the expressivity of certain syntactic goal representations: Markov rewards.

Notably, expressivity is just one kind of question we can ask about the syntax of a goal. Indeed, much of the history on rewards has little to do with its expressivity compared to its computational affordances. Markov rewards are amenable to dynamic programming~\citep{bellman1957markovian}, and therefore fit tidily into a wide variety of existing algorithms for both planning and learning. We suggest that a key set of questions for further research will determine the space of considerations of goals that live on the syntactic and semantic sides of the divide.

\subsubsection{Settling the Reward Hypothesis}

In Abel et al., the bridge from reward to preference come about largely due to the Markov property, when we move beyond the Markov property, the design space of such decisions becomes much richer. 
In subsequent work, \cite{bowling2023settling} expand on the analysis of \cite{abel2021reward} in two ways. First, they go beyond Markovian environments and policies and instead consider the space of partially observable environments \citep{kaelbling1998planning,hutter2000theory,dong2022simple,abel2023crl} and history-based policies. In this case, the agent is only ever exposed to a stream of observations. As such, when we conceive of what the agent wants, these ``wants" must manifest in terms of steering the agent's observable experience. 

Second, Bowling et al. seek a complete characterization of the implicit requirements under which we can relate rewards and preferences. As with the previous work, this analysis demands both compositional and representational commitments.
Here, the compositional commitments are again roughly made in terms of a standard recipe: expected, cumulative, discounted rewards are embraced as the default form of value (though discount is allowed to vary on a per-experience basis, as per \citeauthor{white2017unifying}, \citeyear{white2017unifying}).

On the representational side, Bowling et al. provide two assumptions that create a complete bridge from reward, to value, to the preferences of interest.
First, they assume that \textit{what it means to prefer a policy} $\pi_1$ to another $\pi_2$ is that, eventually, we always prefer the outcomes or states of affairs produced by $\pi_1$ to those of $\pi_2$. This allows intermediate preferences over the outcomes induced by $\pi_1$ and $\pi_2$ to oscillate, possibly for an extended period, so long as they eventually settle into $\pi_1$'s indefinite preference over $\pi_2$. This is one natural view of what it means to prefer one policy to another, but it is again just one choice.
They further assume that a reward expresses a preference over policies just when the values formed by rewards orders \textit{outcomes} in a way that coheres with the policy preferences.
As such, this pair of assumptions tells us both how to connect policy preferences with outcome preferences, and how to connect value to preference.

Then, subject to these two assumptions, Bowling et al. prove the following.

\begin{mdframed}
\begin{theorem}[Informal restatement of Theorem 4.1 by \cite{bowling2023settling}]
    A Markov reward function is capable of representing preference over policies if and only if the preference satisfies the four vNM axioms and a fifth called Temporal $\gamma$-indifference.
\end{theorem}
\vspace{2mm}
\end{mdframed}

We refer the interested reader to the paper and proof by Bowling for full details on the result, and the axioms. For our purposes, we highlight several salient facts. First, we note that questions of expressivity are a meaningful way to understand the richness of different goal languages.
Second, the semantic axioms come in the form of Assumptions 1 and 2 by \cite{bowling2023settling}, which provide a story for connecting value and preference (roughly: policy preferences correspond to finite-horizon outcome preferences). 
However, we suspect this is just the tip of the iceberg. \cite{pires_distr2025}, for instance, expand on this account and explore variations of distributional reinforcement learning \citep{bellemare2017distributional} based on \textit{stock augmented} decision-making that facilitates goals of the kind ``keep the return of this cumulant within this range". A full exploration of possible semantic axioms represents an important direction for future work.

Moreover, while we have here focused our discussion on the particulars of the goal formalisms of reinforcement learning, we suggest that the basic template extends to all goal accounts: that is, goals, by necessity, adhere to the syntax-semantic framing. %

\subsection{Semantic expressivity of different goal languages}

The work reviewed above primarily focuses on the syntax and semantics of goals from the perspective of standard reinforcement learning, with rewards, values, and preferences as the substrate of goals. A complete review of the literature that explores these issues is beyond our scope, but we note a few key lines of work that address questions about goal syntax and semantics. \cite{davidson2025goals}, for instance, ground goals in terms of \textit{reward-producing programs}, a view that is both resonant with Sutton's reward hypothesis, but also adds further structure to how these goals are represented. Other approaches include purely logical forms of goals, as in classical planning \citep{newell1959report} that typically amount to a predicate on states of affairs. Variations of this kind add further richness through incorporating temporal operators, such as linear temporal logic \citep{baier2006planning, littman2017environment, hammond2021multi, richens2025models}, and the reward machines they are often represented by \cite{icarte2022reward}.
Others further incorporate elements of \textit{risk-sensitivity} explicitly into goals \citep{howard1972risk,mihatsch2002risk,lehman2025evolution}. 
Risk-sensitive views are siblings of a broader class of goal accounts that are based in multiple objectives \citep{gabor1998multi}---that is, a decision-maker might have many qualities they care about, and the pursuit of goals is ultimately about balancing these quantities. For example, \cite{ringstrom2023reward} suggests that thirst and hunger might be fundamental, atomic pursuits for decision makers.
Further accounts reject the presence of a discount factor entirely and embrace a truly infinite horizon using the \textit{average reward} \cite{mahadevan1996average,dong2022simple,kumar2023continual}.
Other proposals specifically build around roots in virtue ethics, rather than consequentialism, as articulated by \cite{chang2015value}. Here, Chang argues against the notions of being able to compare any two experiences, as echoed by \cite{aumann1962utility} in his argument against one of the vNM axioms.

\begin{table}[b!]
  \centering
  \caption{
  A non-exhaustive compendium of goal languages from the literature. (Note on goal languages: some are complete, some are partial).
  (DP = Supports dynamic programming; DP- = Does not support dynamic programming; Gen = Facilitates generalization; Prop = Propositional; Markov = Expressivity of Markov reward; Markov+ = Greater expressivity than Markov reward)
  }
  \label{tab:goal-langs}

  \begin{tabular}{llll}
    \toprule
    \makecell[b]{\textbf{Goal} \\ \textbf{Language}} & \makecell[b]{\textbf{Syntactic} \\ \textbf{Affordances}} & \makecell[b]{\textbf{Semantic}\\ \textbf{Expressivity}} \\
    \midrule
    Markov Reward~\citep{sutton2018reinforcement}& DP & Markov \\
    Average Reward~\citep{mahadevan1996average} & DP &  Markov+ \\
    Risk-Sensitive~\citep{howard1972risk} & DP & Markov+ \\
    Distributional~\citep{bellemare2017distributional,bellemare2023distributional} & DP, Gen & Markov \\% \citep{lyle2019comparative} \\
    Constrained \citep{altman2021constrained} & DP & Markov+& \\
    Lexicographic~\citep{shakerinava2025beyond} & DP- & Markov+ \\
    Satisficing \citep{simon1956rational} & Unknown  & Markov+ \\
    Blackwell Optimal \citep{blackwell1962discrete} & DP & Markov+ \\
    First Order Logic~\citep{ghallab1998pddl} & DP, Prop & Markov \\
    Temporal Logic~\citep{icarte2022reward} & DP, Prop  & Markov+, Time, RegEx \\
    Reward Programs~\citep{davidson2025goals} & Diverse & Markov+, Time, Comp \\
    \vdots \\
    \bottomrule
\end{tabular}
\end{table}

As articulated by \cite{davidson2024toward}, different goal representations offer different affordances, even if they ultimately represent the same thing. For instance, two different representations of the same goal can afford different types or degrees of compositionality, generalization, and learning efficiency.
For example, \cite{tasse2020boolean} and \cite{tasse2020task} develop a boolean algebra for \textit{logically} composing goals. Subject to a few minor assumptions, rewards of the usual kind can be cast as members of a boolean algebra, unlocking the operations of conjunction, disjunction, and negation. For example, given the two goals "run the marathon" and "do not get injured", the algebra offers the means of applying a disjunction to these two goal representations directly.
Distributional RL \citep{bellemare2017distributional,bellemare2023distributional} offers another look at representing value, suggesting that decision-makers might benefit by representing the full distribution of future discounted reward. Despite this representation capturing the same effective preference relation (in the sense that taking the expectation yields traditional value), it can still prove useful in generalization \citep{dabney2018implicit}.
We speculate that composition, generalization, and learning efficiency are some of the plausible characteristics that might be desirable of goal representations. For a further examination of goal representations and their trade offs, we refer to \cite{davidson2024toward}.

\subsection{Semantics preserving syntactic operations}%

In language, two syntactically distinct utterances (e.g., ``Rex chased the cat'' versus ``the cat was chased by Rex'') can share the same or similar semantics (i.e., a situation where a dog named Rex is chasing a cat). How might this idea extend to the analysis of goal representations or task representations more generally? Suppose we adopted the account reviewed in the preceding section that formalized goal semantics as \emph{preference orderings over policies} specified by a \emph{value function}. Not only does this clarify what any single goal representation means or represents, it provides a basis for assessing the similarity of different goal representations.

For an example of a semantics-preserving operation for goal representations, consider \emph{potential-based reward shaping}~\citep{ng1999policy}. Given a long-term goal, reward shaping can be viewed as setting intermediate goals that provide short-term motivation. For instance, suppose one has the long-term goal of running the Boston marathon under four hours, but to stay motivated they set the intermediate goal of improving their run time from the day before. A classic result from \cite{ng1999policy} shows that shaping rewards must have a particular functional form (specifically, a difference of potentials) otherwise the optimal behavior under the shaped rewards may differ from that under the original rewards. In our running example, this means that it is not enough to want to improve one's running speed over the previous day; one also needs to be averse to backsliding from the previous day. Crucially, without the additional goal to avoid backsliding, the optimal behavior is to repeatedly improve one's time and backslide, a behavior sometimes described as \emph{exploiting a positive reward cycle}~\citep{ho2019people}. From the perspective of our proposed syntax-semantics distinction, potential-based shaping is a syntactic operation (i.e., a constrained transformation of the original goal representation) that preserves semantic meaning (i.e., the optimal behavior).

Beyond goal representations themselves, work in artificial intelligence and cognitive science also explores how transformations of representations that interact with goals (e.g., world models) can be constrained to preserve optimal behavior. In particular, \cite{grimm2020value} introduce the notion of \emph{value equivalent} representations to explain how deep model-based reinforcement learning systems learn world models that allow effective planning even if they are inaccurate. A similar principle underlies the \emph{value-guided construal} framework, a normative account of how humans form ad hoc world models that trade off value and complexity~\citep{ho2022people,ho2023rational}. To illustrate the idea of world models that induce similar goal semantics, imagine driving to a new destination in a familiar town. Even if you know all the bike paths, you are unlikely to include them in a world model for driving since those details will not change your route (cars cannot be driven on bike paths). Put another way, in the context of driving a car, the simpler model without bike paths preserves the value relations of the more complex model that includes bike paths. From the viewpoint of a syntax-semantics distinction, these value relations are a semantic property, whereas representational complexity is a syntactic property.

Beyond value equivalence and value-guided construal for world models, we propose that drawing a clear analytical distinction between form and meaning can usefully frame research into goal representations. It is thus useful to ask, given some target semantic meaning (e.g., the value function defined by an extrinsic goal) what kinds of syntactic formulations preserve those semantics (i.e., subjective goal representations that induce roughly the same policy ordering). For example, one may have the goal of eating out less, but a semantically similar goal is to cook at home more. The choice of which of these two goal representations may matter: The latter may be a better starting point for planning since it is easier to start thinking about a single dish to cook rather than the many places one could avoid eating out at. More broadly, syntactically distinct but semantically equivalent representations often have different computational affordances. For example, as discussed, Markov rewards enable efficient dynamic programming~\citep{puterman2014markov}. Along similar lines, object-based representations may support more efficient planning while maintaining guarantees about value semantics~\citep{olieslagers2026interactions}.

\section{Discussion}

We have here drawn an analogy between the syntax-semantics divide in studies of language and logic to the way we conceive of goals. Using the standard goal formalisms of reinforcement learning~\citep{sutton2018reinforcement}, this translates to (1) reward as goal atoms, (2) values as compositions of these atoms, and (3) preferences over policies as the representational content referred to by both rewards and values. To move between these three structures, we require additional bedrock commitments that amount to our axioms for goals.
As in language and logic, we consider the syntactic operations that determine how we relate our atoms to sentences---that is, how reward is composed into value. Notably, this composition is often thought to occur both across \textit{time} (in that value incorporates immediate and future reward) and \textit{possibility} (in that value incorporates the likelihood of future rewards).
To relate value to its representational content of a preference relation, we require further commitments that decide how precisely to form this relation in the form of so called semantic axioms. In some cases, this is natural (and arguably so natural that it is taken as implicit): if we embrace vNM \citep{vonneumann1953theory}, then we arrive at an equivalence between value as expected reward and preferences on outcomes. If we further want to impose structure onto our syntax, such as embracing Markovian rewards, then we require additional axioms to ensure this isomorphism is retained \citep{bowling2023settling}.
We saw how several recent accounts in the literature navigated these commitments in order to communicate new results about what kinds of content rewards of different forms can represent. While these accounts center around common formalisms of reinforcement learning, we again speculate that any account of goals that relates to experience will ultimately be expressible in terms of a preference relation of a certain kind. The questions are then: how is that goal represented by an agent (syntax), and what kind of preference does that goal representation refer to (semantics)?

\paragraph{Open Question: Axioms for Goals?}
However, these are just the choices of relations between goal syntax and semantics the community has explored so far. We suggest that this framing highlights a significant conceptual opportunity to explore alternative compositional and semantic accounts of goals.
Moreover, it positions us to scrutinize the needed relationship---if any---across the syntactic and semantic properties of a goal language. If we embrace a particular kind of compositional account of goal, what must this imply, if anything, about our semantic view of goals?
We suggest that exploring the above questions provides a road-map to reaching bedrock clarity on the nature of goals. As summarized by \cref{tab:goal-langs}, we suspect there is a profile of trade-offs in both the syntactic space (compositional affordances, for instance) and the semantic space (expressivity, for instance) that is worth elucidating. %

\paragraph{Open Question: The Starting Point as Syntax or Semantics?}
When conceiving of a goal, the proposal to think of goals in a syntactic or semantic form means that we have a choice: do we begin by committing to a goal as a \textit{syntactic} or a \textit{semantic} object, first? And, how does our theory and analysis need to change depending on our choice of this starting point?
Using the language of rewards and preferences, this amounts to asking what affordances and problems we face if we select the representation as our starting point for thinking about goals, as opposed to the meaning. For example, we might start from rewards, which has arguably been the standard implicit view within reinforcement learning since the dawn of the field. When we start with reward, it offers convenient computational affordances. Specifically, a Markov reward function can be manipulated easily---through approaches like dynamic programming, we find simple mechanisms to compose these rewards into value (subject to suitable syntactic axioms). In fact, the premise that we can easily compute value from reward is so standard that we arguably have chosen our syntactic axioms \textit{in the service of} making this computation straightforward. As an alternative, we could have chosen others from \cref{tab:goal-langs} such as average reward. Yet, such quantities are known for being more difficult to compute than their discounted counterparts. In this way, we frame our syntactic commitments not just in terms of the intrinsic nature of what we can represent, but in terms of the computational properties offered by the syntactic structures we commit to \citep{davidson2024toward}.
If we were to instead start from preferences, we might find that questions about computational complexity or description length enter into our account of how to conceive of preferences in the right way (which has been somewhat absent thus far). We suggest these represent rich directions for further research.

\paragraph{Open Question: Syntactic affordances and semantic expressivity as constraints on cognition?} 
Lastly, a syntax-semantics distinction for goals can provide a useful analytical frame for investigating goal representations in human and animal cognition. In our view, this is not unlike studying the same cognitive process from, say, the computational versus algorithmic levels of analysis~\citep{marr1982vision,anderson1990adaptive}. Specifically, certain properties of goals may be better understood in terms of their syntactic constraints (e.g., how well they afford efficient computation) while others in terms of their semantic constraints (e.g., how well they express rational behaviors in a certain environment). Characterizing how these distinct kinds of constraints interact with one another may also shed light on rational accounts of human cognition~\citep{griffiths2024bayesian} and their computational plausibility~\citep{rich2020intractability}.

\subsection{Conclusion}

Here, we explored the connection between several lines of research on goal representations and the syntax-semantics distinction from linguistics and logic. We propose that delineating the syntactic and semantic properties of goal representations can provide analytical clarity and guide future investigations of the goals that intelligent agents may pursue.

\section*{Acknowledgements}

We would like to thank Logan Walls, Brandon Kaplowitz as well as the reviewers for their valuable feedback on this work. MKH was supported by the National Science Foundation under Grant No. 2434192.

\bibliographystyle{abbrvnat}
\bibliography{goals}

\end{document}